\documentclass[twocolumn]{article}
\usepackage[utf8]{inputenc}
\usepackage{fancyhdr}
\usepackage{authblk}
\usepackage{graphicx}
\usepackage{amsmath}
\usepackage{amssymb}
\usepackage{amsthm}
\usepackage{lipsum}
\usepackage{svg} 
\usepackage{url}

\newtheorem{example}{Example}[section]
\usepackage[english]{babel}
\usepackage[autostyle]{csquotes}
\usepackage{xcolor}
\title{Towards Preserving Semantic Structure in Argumentative Multi-Agent via Abstract Interpretation}

\author{Minal Suresh Patil}

\affil{Umeå Universitet}

\date{}

\begin{document}

\maketitle

\thispagestyle{fancy}
\fancyhead[R]{\textit{Online Handbook of Argumentation for AI, Vol.3}}
\fancyhead[L]{}
\pagestyle{fancy}
\fancyfoot{}

\begin{@twocolumnfalse}
\begin{abstract}
Over the recent twenty years, argumentation has received considerable attention in the fields of knowledge representation, reasoning, and multi-agent systems. However, argumentation in dynamic multi-agent systems encounters the problem of significant arguments generated by agents, which comes at the expense of representational complexity and computational cost. In this work, we aim to investigate the notion of \textit{abstraction} from the model-checking perspective, where several arguments are trying to defend the same position from various points of view, thereby reducing the size of the argumentation framework whilst preserving the semantic flow structure in the system.
\end{abstract}
\end{@twocolumnfalse}

\section{Introduction}
Humans must possess beliefs in order to engage with their surrounding environments successfully, coordinate their activities, and be capable of communicating. Humans sometimes use arguments to influence others to act or realise a particular approach, to reach a reasonable agreement, and to collaborate together to seek the optimal possible solution to a particular problem. In light of this, it is not unexpected that many recent efforts to represent artificially intelligent agents have incorporated arguments and beliefs of their environment. Argumentation-based decision-making approaches are anticipated to be more in line with how people reason, consider possibilities and achieve objectives. This confers particular advantages on argumentation-based techniques, including transparent decision-making and the capability to provide a defensible rationale for outcomes. 

In our recent work, we propose the use of explanations in autonomous pedagogical scenarios~\cite{patil2022explainability} i.e. how explanations should be tailored in multi-agent systems (MAS) (teacher-learner interaction) as shown in Figure 1. It is rational to assume that autonomous agents in open, dynamic, and distributed systems will conform to a linguistic system for expressing their knowledge in terms of one or more ontologies that reflect the salient domain. Agents consequently must agree on the semantics (e.g. privacy) of the terms they use to organize the information, contextualise the environment, and represent different entities in order to engage or cooperate jointly. Abstract argumentation frameworks (AFs)~\cite{dung1995acceptability,bench2007argumentation} are naturally employed for modelling and effectively resolving such types of challenges. In both multi-agent~\cite{maudet2006argumentation} and single-agent~\cite{amgoud2009using} decision-making situations, AFs have been extensively utilised to describe behaviours since they can innately represent and reason with opposing information. Moreover, argumentative models have been presented due to the dialectic nature of AFs so that agents can cooperatively resolve issues or arrive at decisions by communicating implicitly~\cite{dung2009assumption}. 

Present studies of AFs, however, may not be immediately applicable to multi-agent scenarios where agents could come across certain unexpected circumstances in their environment.AFs are naturally used for modelling dynamic systems since, in actuality, the argumentation process is inherently dynamic in nature~\cite{falappa2011evolving, booth2013logical} and this comes with high computational complexity~\cite{dunne2009complexity, dunne2009computational}.

To give a practical example, autonomous Intent-Based Networking (IBN)~\cite{campanella2019intent} captures and translates business intent into network policies that can be automated and applied consistently across the network. The goal is for the network to continuously monitor and adjust its performance to assure the desired business outcome. Intent allows the agent to understand the global utility and the value of its actions. Consequently, the autonomous agents can evaluate situations and potential action strategies rather than being limited to following instructions that human developers have specified in policies. In these cases, agents may adjust their model of the environment as well as their strategy according to information provided by the environment. There are several other circumstances in which the agent may not be able to guarantee a specific status of specific arguments and would necessitate assistance from other agents. Agents may not always know the optimal strategy until they form a coalition. In such circumstances, agents cannot merely compute semantics/conclusions from the ground up since it is not feasible. \enquote{Abstracting} AFs from the original(concrete domain) AF to via Abstract Interpretation can help compute semantics on a much smaller AF. 
Abstraction inherently are necessary if specific properties or specifications of the AF that are abstracted away from them are maintained.

The main contribution of this work is to investigate the semantic properties of the \enquote{abstract} AF from the \enquote{concrete} AF during the multi-agent interactions.
The term \enquote{abstraction} in this work pertains to the notion of abstraction from model checking. Abstraction of the state space may reduce the AF to a manageable size by clustering similar concrete states into abstract states, which can further facilitate verifying these abstract states.
We summarise the primary research question as follows:
\textit{Given a MAS in an uncertain environment, each with a specific subjective evaluation of a given set of conflicting arguments, how can agents reach a consensus whilst preserving specific semantic properties?}   
\begin{figure}[h!]
\centering
\includegraphics[width=1.10\linewidth]{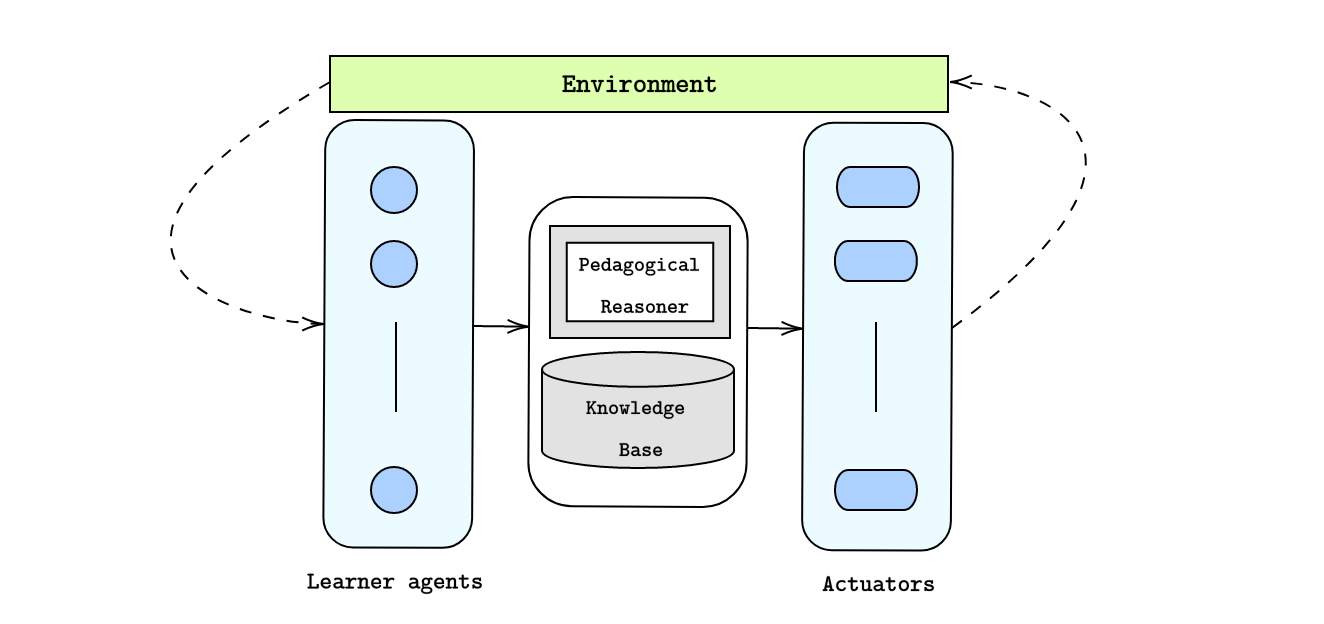}
    \caption{Pedagogical Multi-Agent Reasoning}
    %\captionsetup{justification=centering}
\label{fig:mesh1}
\end{figure}

\section{Method} 

\underline{\textsf{Motivation for Abstract Interpretation:}} 

\noindent Model checking\cite{clarke2000model} is widely accepted as a powerful automatic verification technique for the verification of finite-state systems. Halpern and Vardi proposed the use of model checking as an alternative to the deduction for logics of knowledge~\cite{halpern1991model}. Since then, model checking has been extended to multi-agent systems\cite{hoek2002model}.The state explosion issue is the main impediment to the tractability of model checking. Nevertheless, significant research has been done on this well-known issue, and a variety of approaches have been proposed to circumvent the model checking limitation, including such symbolic methods with binary decision diagrams~\cite{burch1992symbolic}, SAT solvers~\cite{biere1999symbolic}, partial order reduction~\cite{peled1994proving} and abstraction~\cite{clarke2000model}. 

In this work, we focus on abstract interpretation for computing dynamic semantics in MAS. The main point of abstract interpretation\cite{cousot1977abstract} is to replace the formal semantics of a system with an abstract semantics computed over a domain of abstract objects, which describe the properties of the system we are interested in. It formalises formal methods and allows to discuss the guarantees they provide such as soundness (the conclusions about programs are always correct under suitable explicitly stated hypotheses), completeness (all true facts are provable), or incompleteness (showing the limits of applicability of the formal method). Abstract interpretation is mainly applied to design semantics, proof methods, and static analysis of programs. The semantics of programs formally defines all their possible executions at various levels of abstraction. Proof methods can be used to prove (manually or using theorem provers) that the semantics of a program satisfy some specification, that is a property of executions defining what programs are supposed to do. Now, we provide a brief technical primer on key concepts in abstraction interpretation.

\noindent\underline{\textsf{Posets:}}
A partially ordered set (poset) $\langle\mathcal{D}, \sqsubseteq\rangle$ is a set $\mathcal{D}$ equipped with a partial order $\sqsubseteq$ that is (1) reflexive: $\forall x \in \mathcal{D} \cdot x \sqsubseteq x ;(2)$ antisymmetric: $\forall x, y \in \mathcal{D} \cdot((x \sqsubseteq y) \wedge(y \sqsubseteq$ $x)) \Rightarrow(x=y)$; and (3) transitive: $\forall x, y, z \in \mathcal{D} .((x \sqsubseteq y) \wedge(y \sqsubseteq z)) \Rightarrow(x \sqsubseteq z)$. Let $S \in \wp(\mathcal{D})$ be a subset of the poset $\langle\mathcal{D}, \sqsubseteq\rangle$, then the least upper bound \textit{(lub/join)} of $S$ (if any) is denoted as $\sqcup S$ such that $\forall x \in S . x \sqsubseteq \sqcup S$ and $\forall u \in S$. ( $\forall x \in S$. $x \sqsubseteq u) \Rightarrow$ $\sqcup S \sqsubseteq u$, and the greatest lower bound \textit{(glb/meet)} of $S$ (if any) is denoted as $\sqcap S$ such that $\forall x \in S . \sqcap S \sqsubseteq x$ and $\forall l \in S .(\forall x \in S . l \sqsubseteq x) \Rightarrow l \sqsubseteq \sqcap S$. The poset $\mathcal{D}$ has a supremum (or top) $T$ if and only if $T=\sqcup \mathcal{D} \in \mathcal{D}$, and has an infimum (or bottom) $\perp$ iff $\perp=\sqcap \mathcal{D} \in \mathcal{D}$.

\noindent\underline{\textsf{Lattice and Complete Partial Order (CPO):}}
A CPO is a poset $\langle\mathcal{D}, \sqsubseteq, \perp, \sqcup\rangle$ with infimum $\perp$ such that any denumerable ascending chain $\left\{x_i \in \mathcal{D} \mid i \in \mathbb{N}\right\}$ has a least upper bound $\sqcup_{i \in \mathbb{N}} x_i \in \mathcal{D}$. A lattice is a poset $\langle\mathcal{D}, \sqsubseteq, \sqcup, \sqcap\rangle$ such that every pair of elements $x, y$ has a \textit{lub} $x \sqcup y$ and a \textit{glb} $x \sqcap y$ in $\mathcal{D}$, thus every finite subset of $\mathcal{D}$ has a \textit{lub} and \textit{glb}. A complete lattice $\langle\mathcal{D}, \sqsubseteq, \perp, \top, \sqcup, \sqcap\rangle$ is lattice with arbitary subset $S \in \wp(\mathcal{D})$ has a \textit{lub} $\sqcup S$, hence a complete lattice has a supremum $\top=\sqcup \mathcal{D}$ and an infimum $\perp=\sqcup \emptyset$.

\noindent\underline{\textsf{Preorder and Equivalence Relation:}}
A preorder $\preceq$ is a binary relation that is reflexive and transitive, but not necessarily antisymmetric. Then $x \sim y \triangleq x \preceq y \wedge y \preceq x$ is a \textit{equivalence relation} that is reflexive, symmetric $(\forall x, y \in \mathcal{D} . x \sim y \Rightarrow y \sim x)$, and transitive. For any equivalence relation $\sim$, the equivalence class of $x \in \mathcal{D}$ is defined as $[x]_{\sim} \triangleq\{y \in \mathcal{D} \mid y \sim x\}$. The quotient set $\left.\mathcal{D}\right|_{\sim}$ of $\mathcal{D}$ by the equivalence relation $\sim$ is the partition of $\mathcal{D}$ into a set of equivalence classes, i.e. $\left.\mathcal{D}\right|_{\sim} \triangleq\left\{[x]_{\sim} \mid x \in \mathcal{D}\right\}$. Furthermore, the preorder $\preceq$ on $\mathcal{D}$ can be extended to a relation $\preceq \sim$ on the quotient set $\left.\mathcal{D}\right|_{\sim}$ such that $[x]_{\sim} \preceq \sim[y]_{\sim}$ $\Leftrightarrow \exists x^{\prime} \in[x]_{\sim}, y^{\prime}$ $\in[y]_{\sim .} x^{\prime} \preceq y^{\prime}$. Hence, if $\preceq$ is a preorder on $\mathcal{D}$, then $\preceq \sim$ is a partial order on the corresponding quotient set $\left.\mathcal{D}\right|_{\sim}$.

\noindent\underline{\textsf{Abstraction and Galois connection:}}
In the framework for abstract interpretation, Galois connections are used to formalise the correspondence between concrete properties (like sets of traces) and abstract properties (like sets of reachable states), in case there is always a most precise abstract property over-approximating any concrete property. Given two posets $\left\langle\mathcal{D}, \sqsubseteq\right\rangle$  (\textit{concrete domain}) and $\left\langle\mathcal{D}^{\sharp}, \sqsubseteq^{\sharp}\right\rangle$ (\textit{abstract domain}), the pair $\langle\alpha, \gamma\rangle$ of functions $\alpha \in \mathcal{D} \mapsto \mathcal{D}^{\sharp}$ (known as \textit{abstraction function} and $\gamma \in \mathcal{D}^{\sharp} \mapsto \mathcal{D}$ (known as \textit{concretisation function}) forms a \textit{Galois connection} iff $\forall x \in \mathcal{D} . \forall y^{\sharp} \in \mathcal{D}^{\sharp}. \alpha(x) \sqsubseteq^{\sharp} y^{\sharp} \Leftrightarrow x \sqsubseteq \gamma\left(y^{\sharp}\right)$ which is mathematically represented as $\langle\mathcal{D}, \sqsubseteq\rangle \underset{\alpha}{\stackrel{\gamma}{\leftrightarrows}}\left\langle\mathcal{D}^{\sharp}, \sqsubseteq^{\sharp}\right\rangle$ such that (1) $\alpha$ and $\gamma$ are monotonic; (2) $\gamma \circ \alpha$ is extensive (i.e. $\forall x \in \mathcal{D} \cdot x \sqsubseteq \gamma(\alpha(x)))$; (3) $\alpha \circ \gamma$ is reductive (i.e. $\forall y^{\sharp} \in \mathcal{D} \cdot y^{\sharp} \sqsubseteq \alpha\left(\gamma\left(y^{\sharp}\right)\right))$.

\noindent The rationale underpinning Galois connections is that the concrete properties in $\mathcal{D}$ are approximated by abstract properties in $\mathcal{D}^{\sharp}: \alpha(x)$ is the most precise sound over-approximation of $x$ in the abstract domain $\mathcal{D}$ and $\gamma\left(y^{\sharp}\right)$ is the least precise element of $\mathcal{D}$ that can be over-approximated by $y^{\sharp}$. The abstraction of a concrete property $x \in \mathcal{D}$ is said to be \textit{exact} whenever $\gamma(\alpha(x))=x$, in other words, abstraction $\alpha(x)$ of property $x$ loses no information at all. Furthermore, we can say $y^{\sharp} \in \mathcal{D}^{\sharp}$ is a \textit{sound approximation} of $x \in \mathcal{D}$ iff $x \sqsubseteq \gamma\left(y^{\sharp}\right)$.

\section{Discussion}
In this section, we illustrate the nature of \emph{abstraction} in AF and leverage the accrual of arguments whilst preserving the semantic information between them.
\begin{example}
Consider the example provided in ~\cite{nielsen2006generalization}, consisting of the following abstract arguments:
\begin{itemize}
    \item A1: Joe does not like Jack;
    \item A2: There is a nail in Jack’s antique coffee table;
    \item A3: Joe hammered a nail into Jack’s antique coffee table;
    \item A4: Joe plays golf, so Joe has full use of his arms;
    \item A5: Joe has no arms, so Joe cannot use a hammer, so Joe did not hammer a nail into Jack’s antique coffee table.
\end{itemize}
\end{example}

\begin{figure}[h!]
\centering
\includegraphics[width=0.5\textwidth]{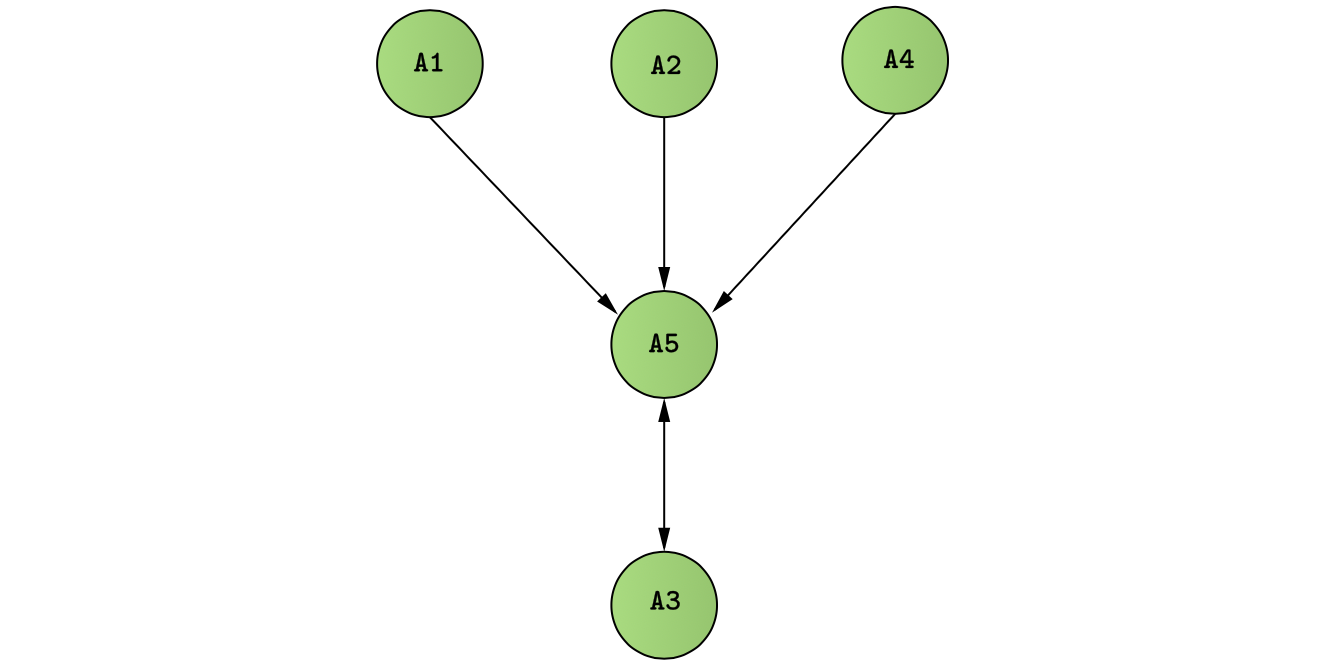}
    \caption{Original AF of Jack and Joe situation}
    %\captionsetup{justification=centering}
\label{fig:mesh1}
\end{figure}

\begin{figure}[h!]
\centering
\includegraphics[width=0.5\textwidth]{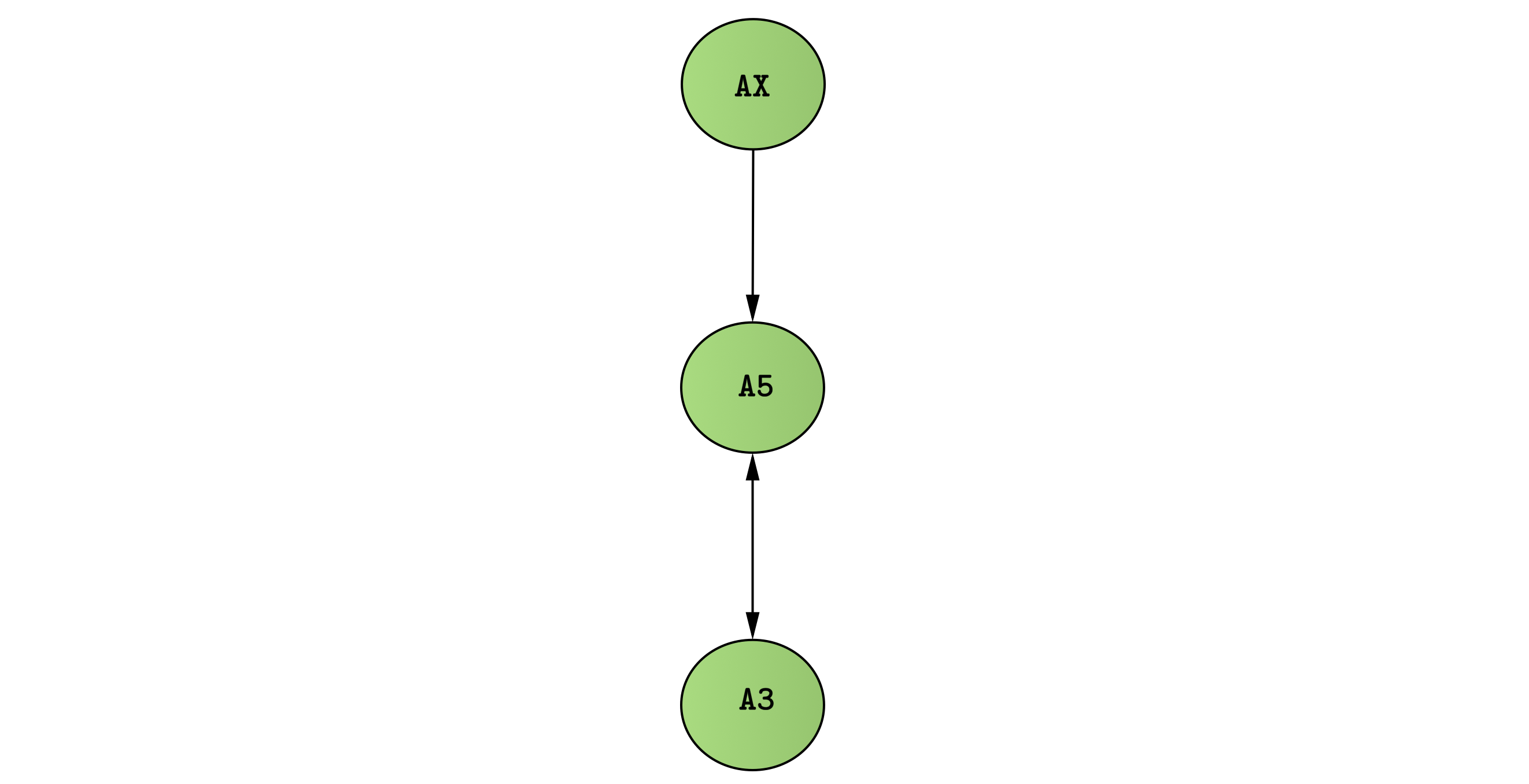}
    \caption{Abstraction of Jack and Joe situation}
    %\captionsetup{justification=centering}
\label{fig:mesh1}
\end{figure}

As we can see in Figure 2, that the argument $A5$ attacks the argument $A3$, whereas arguments $A3$ and $A4$ directly attacks and defeats the argument $A5$.

In our work, employing the abstraction interpretation technique, the semantic relationship between arguments $A1$, $A2$ and $A4$ can be strengthened as $AX$, as shown in Figure 3. Through this simple example, we can reduce the representational complexity of large AFs which further reduces computational cost. This abstraction in multi-agent dynamic AFs can be extended to many realms of argumentation, where auxiliary information (apart from simply winning or losing the argument) come into consideration. One such consideration involves hiding certain information from an opponent e.g. agents abstracting away sensitive and confidential information.

\section{Conclusion}
In this work, we introduced the notion of reducing the complexity of an abstract argumentation framework in a multi-agent setting using abstraction principles from model checking to reduce representational as well as computational cost, which is usually caused due to increased number of arguments in the framework. Furthermore, due to the abstraction of the AF, it would be possible to develop succinct explanations for humans or other agents in the system.

\section*{Acknowledgements}
The author thanks Timotheus Kampik for guidance and valuable insights in this project and the anonymous reviewers for their suggestions and feedback. This work was partially funded by the Knut and Alice Wallenberg Foundation.

%\bibliography{references}
%\bibliographystyle{apalike}

\end{document}